\documentclass[conference]{IEEEtran}
\IEEEoverridecommandlockouts
\usepackage{hyperref}
\usepackage{multirow}

\usepackage{enumitem}
\usepackage{xspace}
\usepackage{algorithm}
\usepackage{algpseudocode}

\usepackage{amsmath,amssymb,amsfonts}
\usepackage{graphicx}
\usepackage{textcomp}
\usepackage{xcolor}
\def\BibTeX{{\rm B\kern-.05em{\sc i\kern-.025em b}\kern-.08em
    T\kern-.1667em\lower.7ex\hbox{E}\kern-.125emX}}
\begin{document}

\title{CFCPalsy: Facial Image Synthesis with Cross-Fusion Cycle Diffusion Model for Facial Paralysis Individuals
}

\author{
    \IEEEauthorblockN{Weixiang Gao, Yating Zhang, Yifan Xia$^{\dagger}$}
    \IEEEauthorblockA{
        School of Mechanical, Electrical \& Information Engineering\\
        Shandong University, Weihai, China\\
        gaoweixiang@mail.sdu.edu.cn, yatingzhang@mail.sdu.edu.cn, xiayifan@sdu.edu.cn
    }
}


\maketitle

\renewcommand{\thefootnote}{\fnsymbol{footnote}}
\footnotetext[2]{Corresponding author} 

\begin{abstract}
Currently, the diagnosis of facial paralysis remains a challenging task, often relying heavily on the subjective judgment and experience of clinicians, which can introduce variability and uncertainty in the assessment process. One promising application in real-life situations is the automatic estimation of facial paralysis. However, the scarcity of facial paralysis datasets limits the development of robust machine learning models for automated diagnosis and therapeutic interventions. To this end, this study aims to synthesize a high-quality facial paralysis dataset to address this gap, enabling more accurate and efficient algorithm training. Specifically, a novel Cross-Fusion Cycle Palsy Expression Generative Model (CFCPalsy) based on the diffusion model is proposed to combine different features of facial information and enhance the visual details of facial appearance and texture in facial regions, thus creating synthetic facial images that accurately represent various degrees and types of facial paralysis. We have qualitatively and quantitatively evaluated the proposed method on the commonly used public clinical datasets of facial paralysis to demonstrate its effectiveness. Experimental results indicate that the proposed method surpasses state-of-the-art methods, generating more realistic facial images and maintaining identity consistency. {\hypersetup{urlcolor=red}\href{https://github.com/GaoVix/CFCPalsy}{\textcolor{blue}{Code Link}}}

\end{abstract}

\begin{IEEEkeywords}
Diffusion Models, Facial Image Synthesis, Deep Learning, Facial Paralysis Images, Expression Transfer
\end{IEEEkeywords}

\section{Introduction}
\label{sec:intro}
Facial palsy, which can result in temporary or permanent weakness or paralysis of one side of the face, affects approximately 23 out of every 100,000 people annually \cite{ge2022automatic}. Beyond the physical discomfort, patients with facial palsy often experience significant psychological distress \cite{hotton2020psychosocial}. Accurate and timely diagnosis is crucial for effective treatment; however, the variability in clinical presentation poses significant challenges, and the severity estimation of facial palsy remains a largely subjective process \cite{ge2022automatic}. 

\begin{figure}[tb] \centering
    \includegraphics[width=0.48\textwidth]{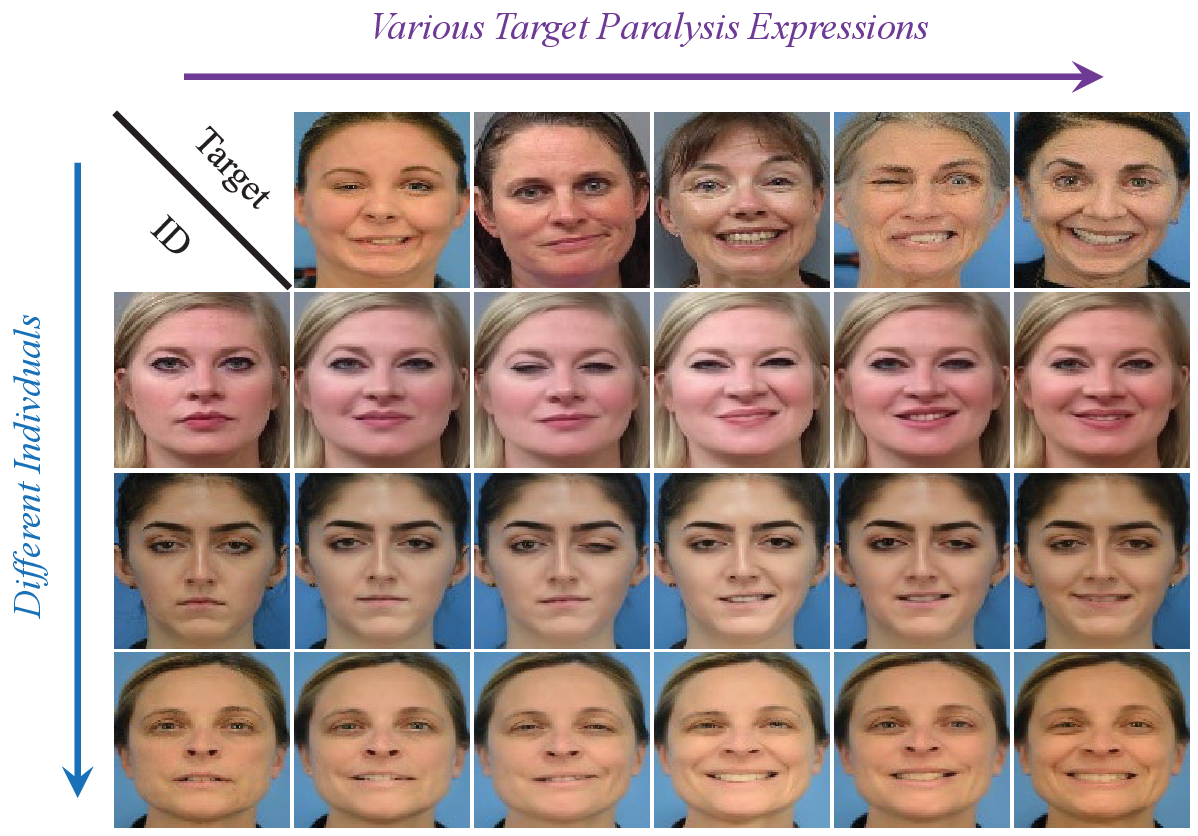}
    \caption{Facial palsy synthesis results. Each row contains images from the same individual, and each column shows images with the same facial palsy expression.} \label{fig:figure1}
\end{figure}

In recent years, machine learning (ML) has shown promise in aiding the diagnosis of facial paralysis by analyzing facial features and expressions \cite{vrochidou2023automatic,amsalam2024facial,gaber2022classification,ge2022automatic}. However, the development of reliable ML models is limited by the lack of comprehensive, annotated datasets representing the diversity and severity of facial paralysis. Collecting high-quality images of facial palsy patients is challenging due to privacy concerns and the sensitive nature of medical data. Existing datasets \cite{AFLFP,MEEI,YFP} are often limited in size, scope, and variability, restricting the generalization ability and performance of ML algorithms.

Several studies have explored the use of generative adversarial networks (GANs) \cite{GAN} to generate facial paralysis face images \cite{yaotome2019simulation,sakai2021simulation,sajid2018automatic}. Although \cite{yaotome2019simulation} effectively employs GANs to simulate facial expressions by reconstructing images from boundary data, the generated images often lack the fine-grained details required for accurate medical diagnosis, especially when significant differences exist between the patient's facial features and the target face incorporating palsy characteristics. Building on this, \cite{sakai2021simulation} incorporates facial shape normalization to mitigate unnatural appearances caused by shape differences between patients and models. Although this approach enhances image realism, it remains limited in capturing intricate facial expressions and fine details, reducing its utility in nuanced diagnostic cases. In addition to employing data augmentation, GANs are utilized to enhance CNN performance in grading facial palsy \cite{sajid2018automatic}. By generating face images with varying degrees of palsy, this method mitigates overfitting and improves model generalization. However, the synthesized images are limited to five categories, failing to capture the full spectrum of facial expressions and subtle asymmetries observed in real-world cases.

This paper addresses the critical need for a comprehensive facial paralysis dataset by synthesizing high-quality images. Our generator is designed to capture diverse paralysis conditions, producing face images with varying severity levels based on a given facial identity. This dataset serves as a valuable resource for training ML models, with the potential to enhance diagnostic accuracy and improve clinical outcomes.

To this end, we propose the Cross-Fusion Cycle Palsy Expression Generative Model (CFCPalsy), a novel triple-condition image generation framework. CFCPalsy generates realistic facial images by combining the identity of a given ID image with the expression and landmark features of a style image. By leveraging landmark features and a corresponding loss function, the model effectively captures nuanced facial deformations, including detailed changes in features such as the eyes, mouth, and overall asymmetry, ensuring high fidelity in depicting facial palsy characteristics. When fed multiple features simultaneously, models often face challenges like feature redundancy and conflicting information, leading to degraded performance \cite{li2017feature}. Redundancy can cause the model to overweight certain features, while conflicts result in inconsistent learning and predictions. To overcome these issues, CFCPalsy employs a novel cross-fusion feature integration method. This approach harmonizes information from multiple features, leveraging their complementary strengths to enhance the model's performance and robustness significantly. Training diffusion models \cite{ddpm,diffusion} on limited datasets often results in inefficiencies and suboptimal performance due to data scarcity. To address this, we enhanced training efficiency and performance under constrained data conditions by introducing a modified training strategy. Our approach includes the cycle diffusion framework, which optimizes the learning process and enables the model to deliver superior results even with limited data.


\vspace{3pt}
On the whole, the contributions of this paper are summarized as follows:  
\begin{itemize}[itemsep=0pt,parsep=0pt,topsep=2bp]
    \item We first combine diffusion models with landmark features to improve the quality of facial synthesis and introduce a clever feature confusion strategy that substantially improves the model's capability to effectively learn from various features. 
    \item We propose a cycle framework for diffusion models, which provides an innovative approach to train higher-performance models with limited datasets.
    \item Our approach introduces a novel application of generative models for facial paralysis synthesis, achieving state-of-the-art results and addressing the scarcity of medical datasets, thereby facilitating future research in automated facial analysis.
\end{itemize}


\begin{figure*}[tb] \centering

    \includegraphics[width=\textwidth]{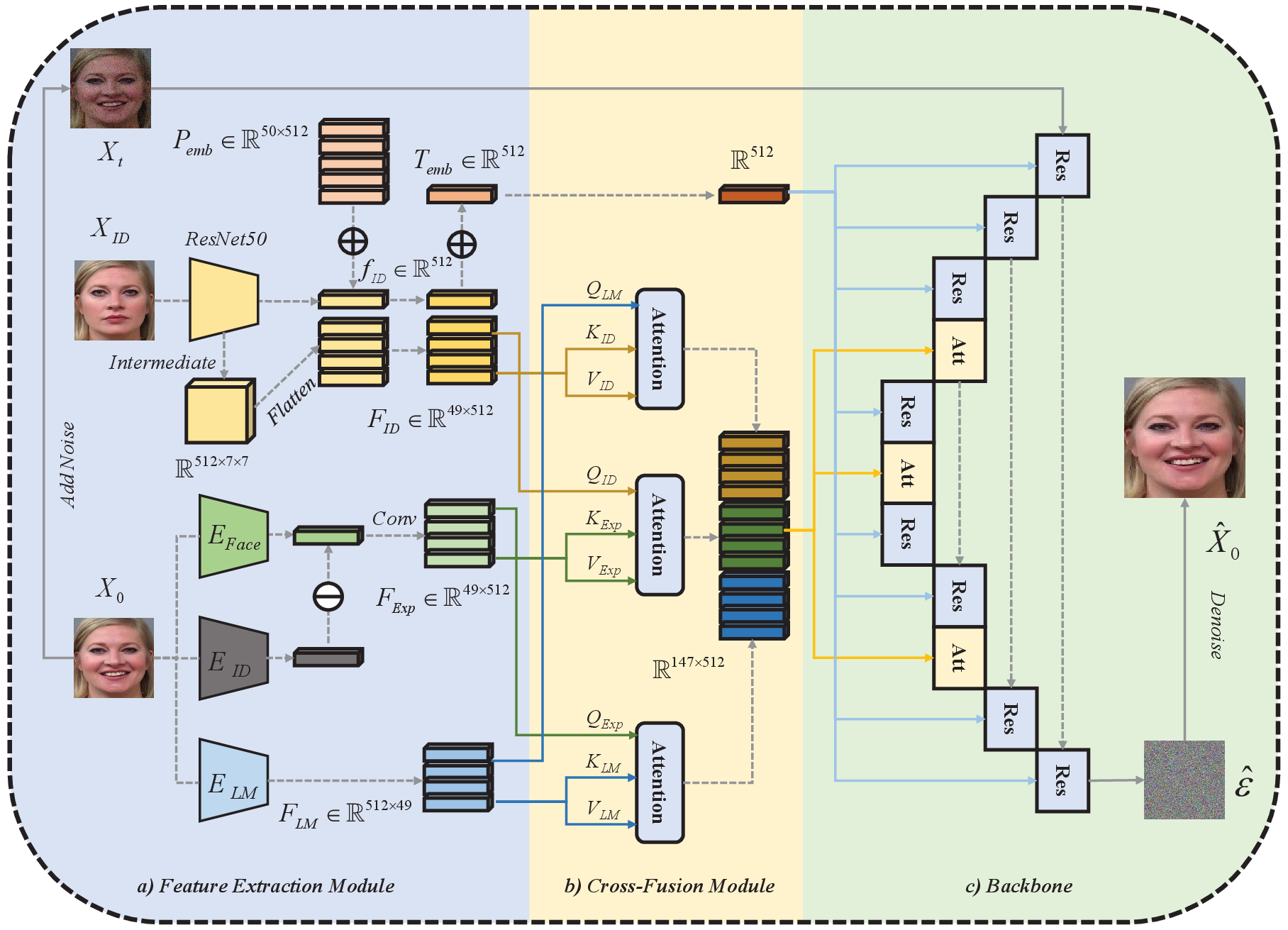}
    \caption{An overview of the forward process of CFCPalsy. a) The architecture of our feature extractors. We train the model using different images of the same patient, with one serving as the identity image $X_{ID}$ and the other as the facial palsy expression image $X_{0}$. The identity model extracts identity features from the identity image. The expression module (including $E_{Face}$ and $E_{ID}$) and $E_{LM}$ extract expression and landmark features from the facial palsy expression image, respectively. b) An illustration of the principle of the cross feature fusion strategy. We employ a method similar to cross-attention to facilitate information exchange among the three conditional features. After this interaction, the features are concatenated for further processing. c) A diagram of the noise predicting network. CFCPalsy utilizes a classic U-Net architecture combined with residual connections and cross-attention mechanisms to accurately predict the noise.} \label{fig:figure2}
\end{figure*}

\section{Related Work}
\label{sec:related_works}
\paragraph{Face Synthesis} Face generation has become a prominent research area in computer vision, driven by advancements in generative models such as Generative Adversarial Networks (GANs) \cite{GAN}, Variational Autoencoders (VAEs) \cite{vae3,vae4}, and diffusion models \cite{ddpm,diffusion}. GANs have been widely applied to tasks like face generation and augmentation \cite{pumarola2018ganimation,sun2019single,xiao2018elegant}, with methods developed to retain original identity while enabling modifications such as image styles or backgrounds \cite{qiu2021synface,deng2020disentangled,bae2023digiface}. Recently, researchers have turned to diffusion models \cite{huang2023collaborative,zeng2023face,stan2023facediffuser}, which provide a more stable and probabilistic framework for face image generation. For instance, DCFace \cite{kim2023dcface} introduces a dual-condition diffusion model to generate synthetic face datasets with varied identities and styles. Building on this, our work explores how the severity of facial palsy and changes in facial expressions impact the synthesis of realistic images.

\paragraph{Diffusion Models}
Diffusion models \cite{ddpm,diffusion} have emerged as a powerful alternative to GANs for image generation tasks \cite{dhariwal2021diffusion}. Unlike GANs, which rely on adversarial training, diffusion models iteratively refine noisy inputs, producing stable and high-fidelity outputs. Recent advancements include enhanced sampling strategies \cite{ho2022classifier,salimans2022progressive,karras2022elucidating} and latent space diffusion techniques \cite{ergasti2024towards,podell2023sdxl,mitchel2024single}, which improve efficiency and output quality. Additionally, studies \cite{kim2024improving,zheng2023fast,karras2024analyzing} have optimized training efficiency through modified strategies. Building on this, we propose a novel cycle diffusion strategy incorporating a secondary loss function to improve performance, particularly on limited datasets.


\section{Method}
In this section, we introduce the Cross-Fusion Cycle Palsy Expression Model (CFCPalsy) for synthesizing high-quality facial paralysis images (see Fig. \ref{fig:figure2}). Section \ref{sub:3.1} outlines the foundational concepts of diffusion models. In Section \ref{sub:3.2}, we describe the feature extractors integrated into CFCPalsy, which enhance the model's ability to capture fine facial details by accurately extracting diverse features. Section \ref{sub:3.3} details the proposed cross-feature fusion mechanism, which reduces over-reliance on individual features by exchanging query vectors derived from different features, thereby improving overall learning capability. Lastly, Section \ref{sub:3.4} introduces a cycle training strategy that boosts training efficiency and performance on limited datasets through process modifications and the inclusion of a secondary loss function.

\subsection{Preliminary}
\label{sub:3.1}

Diffusion models \cite{ddpm,diffusion} are a class of generative models that define a forward process, where data is gradually corrupted by noise, and a reverse process, which learns to denoise and recover the original data distribution. The forward process is typically modeled by a discrete-time Markov chain \cite{markov1971extension}, where at each time step \( t \), the data \( X_t \) is obtained by adding Gaussian noise to the data from the previous step \( X_{t-1} \), following the equation:
\begin{equation}
    q(X_t \mid X_{t-1}) = \mathcal{N}(X_t; \sqrt{1-\beta_t} \, X_{t-1}, \beta_t I), 
    \label{eq:1}
\end{equation}
where \( \beta_t \) is a variance schedule that controls the amount of noise added at each step. The reverse process aims to estimate the posterior distribution \( p_\theta(X_{t-1} \mid X_t) \), which is modeled by a neural network parameterized by \( \theta \). The training objective is to minimize the variational bound on the negative log-likelihood, which often reduces to a simplified mean squared error loss between the true noise and the noise predicted by the model:
\begin{equation}
    L (\theta) = \mathbb{E}_{t, X_t, \epsilon}\left[\|\epsilon - \epsilon_\theta(X_t, t)\|^2\right],
    \label{eq:2}
\end{equation}
where \( \epsilon \) is the true noise and \( \epsilon_\theta(X_t, t) \) is the model’s prediction. By iteratively applying this reverse process, the model is able to generate new data samples from pure noise, refining them step by step until the desired output is obtained. Building on this foundation, we employ a two-stage diffusion process along with a more complex loss function to enhance training efficiency and improve the quality of the generated outputs.

\subsection{Feature Extraction Module}
\label{sub:3.2}
\paragraph{ID Extraction Module}To accurately capture identity information from $X_{ID}$, we use the IR-50 \cite{deng2019arcface}, a typical CNN, as the identity feature extractor. We utilize both the output features $f_{ID} \in \mathbb{R}^{512}$ from the final layer and the intermediate layer features $I_{ID} \in \mathbb{R}^{512 \times 7 \times 7} $of the model to enhance the quality of image generation. For $I_{ID}$, we first flatten and rearrange their dimensions to obtain $F_{ID} \in \mathbb{R}^{49 \times 512}$ that facilitates subsequent processing. Then, we concatenate the two sets of features and add positional embeddings $P_{emb}$. In summary, the output of the identity feature extraction module $M_{ID}$ is formulated as follows:
\begin{equation}
    M_{ID}(X_{ID}) = Concat[f_{ID}, Flatten(I_{ID})] + P_{emb}.
    \label{eq:3}
\end{equation}
For the output features, we process the intermediate feature part and the final layer feature part separately by injecting them into different model blocks, rather than concatenating them together. The identity feature extractor is trainable during the diffusion model training process to effectively extract identity features.

\paragraph{Expression Extraction Module}CFCPalsy employs the deviation framework of DLN \cite{zhang2021learning} to accurately capture the expression features. This module $M_{Exp}$ consists of two models: a facial feature extraction model $E_{Face}$ and an identity feature extraction model $E_{ID}$, sharing the same network structure. Based on the assumption that facial features can be divided into identity features and expression features, we use two separate models to extract expression and identity features from the same facial paralysis image $X_0$. By subtracting the identity features from the facial features, we obtain expression features that are invariant to identity. The subtracted features are then passed through a convolution operation for channel alignment. Specifically, 
\begin{equation}
    M_{Exp}(X_{0}) = Conv[E_{Face}(X_0)-E_{ID}(X_0)].
    \label{eq:4}
\end{equation}

\paragraph{Landmark Extractor}
Facial landmarks refer to key points on a face such as the corners of the eyes, the tip of the nose, and the edges of the lips that are used to represent and understand facial geometry, which could enhance facial generation by imposing constraints on critical facial regions. For the first time, we integrate facial landmark features with diffusion models, coupled with a specifically designed loss function, to generate realistic and detailed facial images of patients with facial palsy. In our CFCPalsy, we employ the MobileFaceNet \cite{chen2021pytorch} as the landmark detector. We remove its output layer to use it as a landmark feature extractor $ E_{LM} $, which outputs a feature $ F_{LM} \in \mathbb{R}^{49\times 512} $.

\subsection{Cross-Fusion Module}
\label{sub:3.3}
The attention mechanism \cite{vaswani2017attention} has become essential in many state-of-the-art models, particularly for tasks requiring integration across data sequences or sources. By dynamically weighting elements within a sequence, it focuses on the most relevant parts to enhance feature extraction and decision-making. Extending this, the cross-attention mechanism operates between two sequences, enabling effective fusion of information from multiple modalities or input streams.

We aim to enable the model to fully leverage identity, expression, and landmark information when predicting noise. While directly concatenating features from these sources and injecting them into the model is a straightforward approach, it often leads to unsatisfactory results—generated images either lose identity details from $ X_{ID} $ or fail to preserve the facial expression features from $ X_{0} $. To address this, we adopt a cross-attention mechanism, allowing mutual interaction among features. Here, the query sequence of one feature guides the extraction of information from the key-value sequence of another, ensuring a more effective integration of features.

The features from the extractors are first mapped into the query, key, and value vector spaces separately through a QKV encoder. Specifically, 
\begin{equation}
    Q = FW^Q, K = FW^K, V = FW^V,
    \label{eq:7}
\end{equation}
where $F$ indicates the three types of features extracted from the former modules, and $ W^Q,W^K,W^V $ represent the model's learnable parameter matrixes. Similar to the multi-head attention mechanism \cite{vaswani2017attention}, we utilize multiple sets of transformation matrices to obtain multiple vector spaces, aiming to capture different interaction patterns among features. 

Subsequently, the identity features, expression features, and landmark features each have their own corresponding QKV vector spaces. Next, we perform vector interactions. Specifically, 
\begin{equation}
    \text{Attention}_{ID} = \text{Softmax}\left(\frac{Q_{LM} K_{ID}^T}{\sqrt{d_k}}\right)V_{ID},
    \label{eq:a1}
\end{equation}
\begin{equation}
    \text{Attention}_{Exp} = \text{Softmax}\left(\frac{Q_{ID} K_{Exp}^T}{\sqrt{d_k}}\right)V_{Exp},
    \label{eq:a2}
\end{equation}
\begin{equation}
    \text{Attention}_{LM} = \text{Softmax}\left(\frac{Q_{Exp} K_{LM}^T}{\sqrt{d_k}}\right)V_{LM},
    \label{eq:a3}
\end{equation}
where $d_k$ is a scaling factor.

\begin{figure}[!tb] \centering
    \includegraphics[width=0.48\textwidth]{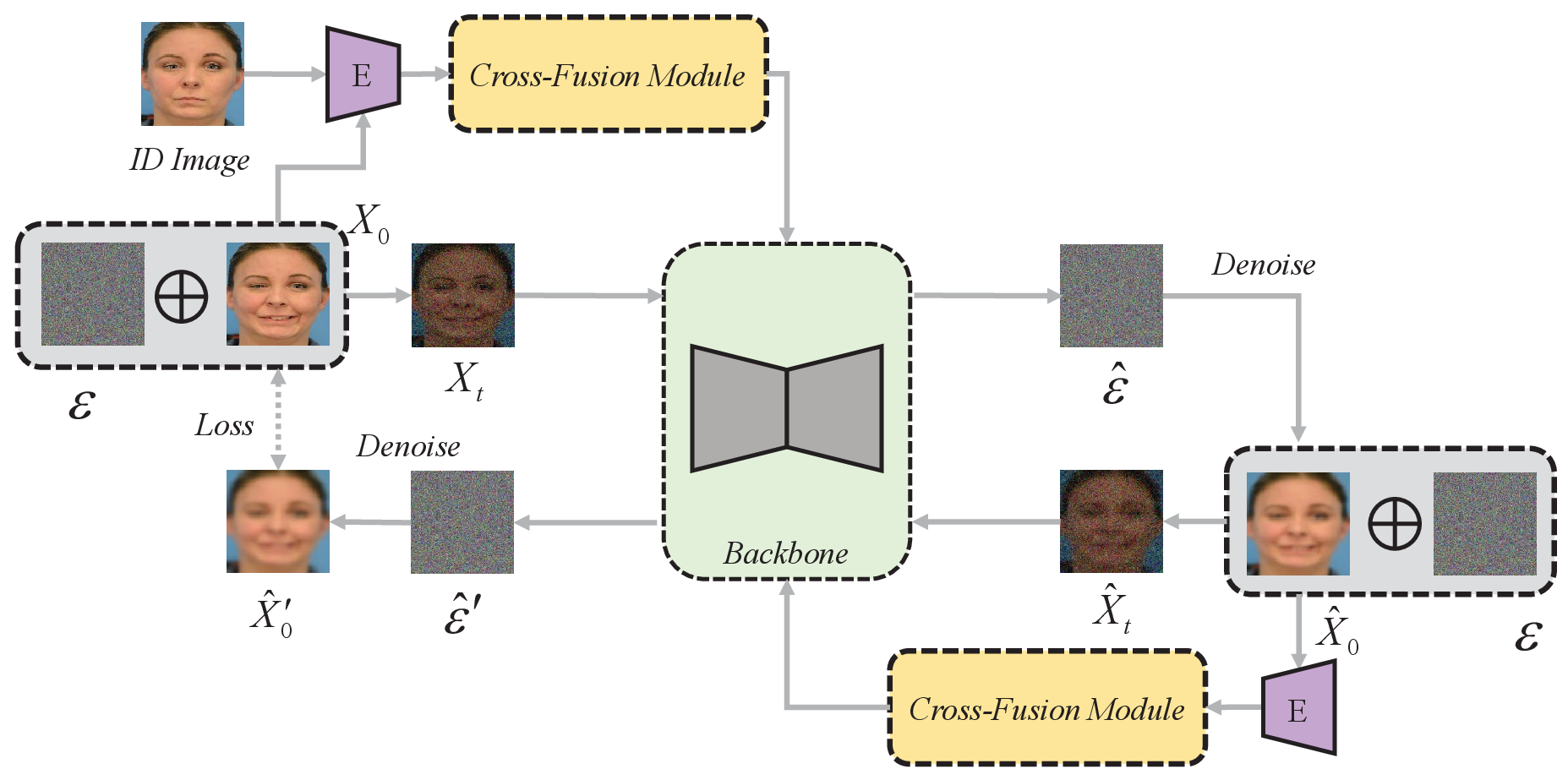}
    \caption{An illustration of the cycle training strategy of CFCPalsy. Each training data undergoes two diffusion processes, where the plus sign indicates the noise addition process.} \label{fig:loss}
\end{figure}

\subsection{Cycle Diffusion Strategy}
\label{sub:3.4}

Building upon the training framework of the original diffusion model, we incorporated an additional step of secondary diffusion. Algorithm \ref{alg:diffusion_training} shows the training procedure of CFCPalsy. After the original diffusion process, where the input image $ X_0 $ is noised to generate a noisy image $ X_t $, and the network predicts the noise $ \epsilon_\theta $ to restore the original image $ \hat{X_0} $. Based on the equation
\begin{equation}
    \hat{X_0} = \frac{1}{\sqrt{\bar{\alpha}_t}} \left( X_t - \sqrt{1 - \bar{\alpha}_t} \epsilon_\theta(X_t, t) \right),
    \label{eq:5}
\end{equation}
where \(\bar{\alpha}_t\) is a term that accumulates the effects of the noise schedule over time, often defined as the product \(\bar{\alpha}_t = \prod_{s=1}^{t} \alpha_s\), with \(\alpha_s\) being a scalar controlling the amount of noise added at each time step,
we further introduced a secondary diffusion step (see Fig. \ref{fig:loss}). In this step, the predicted image $ \hat{X_0} $ is treated as the new input image, which is subjected to the same noise addition process to obtain a secondary noisy image $ \hat{X_t} $. This secondary noisy image is then fed into the noise prediction network to generate a secondary predicted noise $ \epsilon^{'}_\theta $, which is subsequently used to restore the once-noised image.

For the first diffusion process, we utilized a weighted combination of mean square error loss (see Eq. \ref{eq:2}), identity consistency loss $L_{ID}$ (the same as the Time-step Dependent ID Loss in DCFace \cite{kim2023dcface}), and landmark loss $L_{LM}$ as our first loss function. The identity consistency loss $L_{ID}$ is derived from the cosine similarity between the identity features extracted from the images, while the landmark loss $L_{LM}$ is obtained by calculating the mean squared error of the detected landmark keypoints:
\begin{equation}
    L_{first} = L(\theta) + \lambda L_{ID}(X_{ID}, \hat{X_0}) + \gamma L_{LM}(X_{0}, \hat{X_0}),
    \label{eq:loss1}
\end{equation}

To align with the architecture of CFCPalsy, we introduced the concept of a secondary loss function to enhance the effectiveness of the model's training, whose basic composition is consistent with that of the first loss. However, the noisy image $ \hat{X_t} $ generated from $ \hat{X_0} $ does not participate in the denoising process after the second diffusion, and $ \hat{X_0} $ itself is also excluded from the calculation of the second loss. Instead, we calculate the respective loss components using the original image $ X_0 $ before the first diffusion and the restored predicted image $\hat{X_0}^{'}$ after the second denoising. Specifically,
\begin{equation}
    L_{second} = L(\theta') + \lambda L_{ID}(X_{ID}, \hat{X_0'}) + \gamma L_{LM}(X_{0}, \hat{X_0'}),
    \label{eq:loss2}
\end{equation}
The final loss is the weighted sum of the first loss and the second loss.

\section{Experiments}
\label{sec:Experiments}
\subsection{Datasets}
\paragraph{AFLFP} The Annotated Facial Landmarks for Facial Palsy (AFLFP) \cite{AFLFP} dataset is a comprehensive, laboratory-controlled collection designed for the analysis of facial paralysis. It includes facial images from 88 subjects, each participating in 16 expression videos, such as brow raise, closed smile, gentle eye closure, and open smile. For each expression, the dataset captures keyframes corresponding to four critical states: neutral, onset, mid-state (between onset and peak), and peak. In total, the AFLFP dataset contains 5,632 facial images, with 1,408 images for each key state and 64 images per subject.


\paragraph{MEEI} The Massachusetts Eye and Ear Infirmary (MEEI) \cite{MEEI} dataset is a curated collection focused on facial palsy, comprising videos and high-resolution images. It includes 60 videos from 9 healthy subjects and 51 patients with varying degrees of facial paralysis. Each video captures 8 distinct facial movements, which are also provided as individual images, resulting in a total of 480 high-resolution images. Subjects are categorized based on their eFACE \cite{banks2015clinician} scores, which assess the severity of facial palsy.


\begin{algorithm}[tb]
    \caption{Training Algorithm of CFCPalsy}
    \label{alg:diffusion_training}
    \begin{algorithmic}[0]
        \For{every batch}
            \For{$(x_0, x_{id})$ in the batch}
            \State Sample $ \epsilon \sim \mathcal{N}(0, I)$
            \State Sample $ t \sim \text{Uniform}(\{0, 1, 2, \dots, T\})$
            \State Perform initial diffusion
            \State $ x_t = \sqrt{\bar{\alpha}_t}x_0 + \sqrt{1 - \bar{\alpha}_t}\epsilon$ 
            \State $\hat{x}_0 = \frac{1}{\sqrt{\bar{\alpha}_t}} \left( x_t - \sqrt{1 - \bar{\alpha}_t} \epsilon_\theta(x_t, x_{id}, t) \right)$
            \State Calculate $\mathcal{L}_{first}$
            \State Perform second diffusion
            \State $ x_t^{'} = \sqrt{\bar{\alpha}_t}\hat{x}_0 + \sqrt{1 - \bar{\alpha}_t}\epsilon$ 
            \State $\hat{x}^{'}_0 = \frac{1}{\sqrt{\bar{\alpha}_t}} \left( x_t - \sqrt{1 - \bar{\alpha}_t} \epsilon_\theta^{'}(x_t^{'}, x_{id}, t) \right)$
            \State Calculate $\mathcal{L}_{second}$
            \State Update with $\mathcal{L}=\mathcal{L}_{first} + \sigma \mathcal{L}_{second}$
            \EndFor
        \EndFor
    \end{algorithmic}
\end{algorithm}

\subsection{Evaluation Metrics}

\begin{figure}[!tb] \centering
    \includegraphics[width=0.48\textwidth]{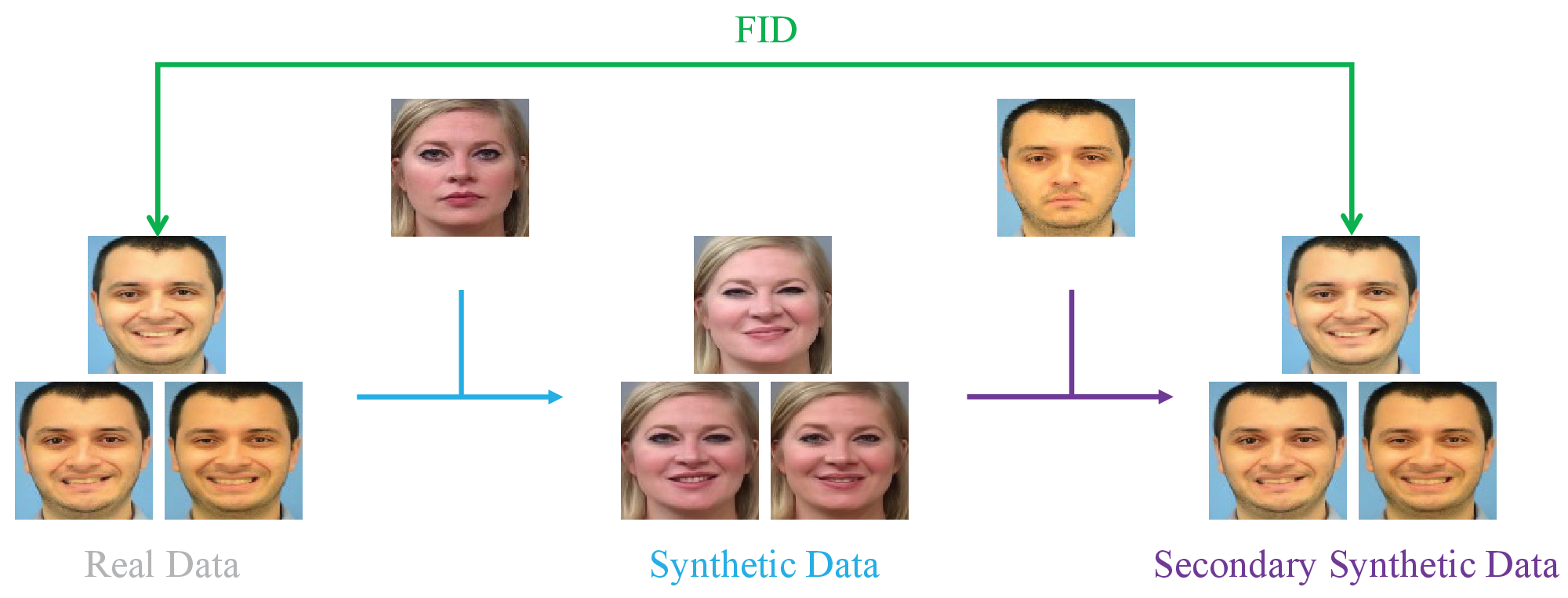}
    \caption{A schematic diagram of the aFID calculation method. Calculate the FID value between the twice-synthesized data and the original data.} \label{fig:fid}
\end{figure}

\begin{figure*}[!ht] \centering
    \includegraphics[width=0.94\textwidth]{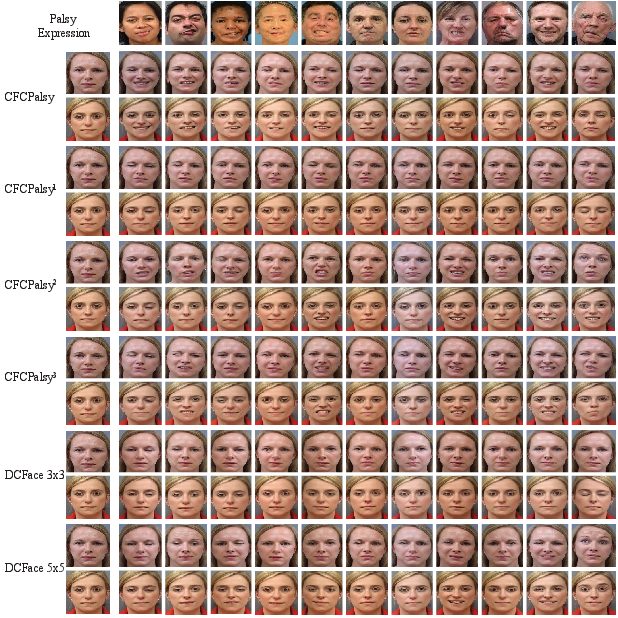}
    \caption{Visualized experimental results. CFCPalsy is the standard version, CFCPalsy$^1$ is the CFCPalsy without facial landmarks, CFCPalsy$^2$ excludes the cycle training strategy and CFCPalsy$^3$ operates in the absence of the cross-fusion module.} \label{fig:vis}
\end{figure*}

To evaluate the effectiveness of image generation from different perspectives, we compute Peak Signal-to-Noise Ratio (PSNR), Structural Similarity Index Measure (SSIM) \cite{wang2004image}, Learned Perceptual Image Patch Similarity (LPIPS) \cite{zhang2018perceptual}, and Perceptual Loss (PL) \cite{johnson2016perceptual} on unseen data. For this multi-condition synthesis task, these metrics are computed separately for identity images and facial paralysis expression images, allowing simultaneous evaluation of identity preservation and expression transfer.

Fr\'{e}chet Inception Distance (FID) \cite{heusel2017gans} is a widely used metric for evaluating image generation quality. It quantifies the distance between the distributions of generated and real images by comparing feature statistics extracted from a pre-trained Inception network. A lower FID score indicates greater similarity between the generated and real images in terms of visual quality and diversity. However, some studies \cite{chen2023examining,qin2023class} have shown that inconsistencies in class distributions between the generated and real datasets can affect the FID score calculation.

\begin{table*}[!htb]

\caption{Quantitative results on the test set.}
\label{tab:metric}
\vspace{1em}
\centering
\begin{tabular}{ccccc|ccccccccc}
\hline
\multicolumn{5}{c|}{Method} & \multicolumn{9}{c}{Metrics} \\ \hline
\multirow{2}{*}{Model} & \multirow{2}{*}{GS} & \multirow{2}{*}{LM} & \multirow{2}{*}{CF} & \multirow{2}{*}{CD} & \multirow{2}{*}{aFID $\downarrow$} & \multicolumn{2}{c}{PSNR $\uparrow$} & \multicolumn{2}{c}{SSIM $\uparrow$} & \multicolumn{2}{c}{LPIPS $\downarrow$} & \multicolumn{2}{c}{PL $\downarrow$} \\
 &  &  &  &  &  & \multicolumn{1}{l}{X$_{ID}$} & \multicolumn{1}{l}{X$_{Style}$} & \multicolumn{1}{l}{X$_{ID}$} & \multicolumn{1}{l}{X$_{Style}$} & \multicolumn{1}{l}{X$_{ID}$} & \multicolumn{1}{l}{X$_{Style}$} & \multicolumn{1}{l}{X$_{ID}$} & \multicolumn{1}{l}{X$_{Style}$} \\ \hline
\multirow{2}{*}{DCFace \cite{kim2023dcface}} & 3x3 & - & - & - & 56.77 & 15.95 & 17.07 & 0.32 & 0.35 & 0.37 & 0.34 & 0.27 & 0.24 \\
 & 5x5 & - & - & - & 56.20 & 17.15 & \textbf{19.39} & 0.40 & \textbf{0.48} & 0.34 & \textbf{0.26} & 0.26 & \textbf{0.19} \\ \hline
\multirow{4}{*}{CFCPalsy} & - & $\times$ & $\checkmark$ & $\checkmark$ & 50.47 & 17.31 & 16.05 & 0.45 & 0.40 & \textbf{0.31} & 0.33 & \textbf{0.25} & 0.28 \\
 & - & $\checkmark$ & $\times$ & $\checkmark$ & 54.42 & 17.13 & 15.99 & \textbf{0.47} & 0.44 & \textbf{0.31} & 0.33 & 0.26 & 0.26 \\
 & - & $\checkmark$ & $\checkmark$ & $\times$ & 45.69 & 17.41 & 16.31 & \textbf{0.47} & 0.43 & 0.32 & 0.34 & \textbf{0.25} & 0.28 \\
 & - & $\checkmark$ & $\checkmark$ & $\checkmark$ & \textbf{43.76} & \textbf{17.44} & 16.36 & \textbf{0.47} & 0.45 & \textbf{0.31} & 0.33 & \textbf{0.25} & 0.27 \\ \hline
\end{tabular}
\vspace{1em} \\
\centering
\begin{tabular}{l}
\textbf{GS} = Grid Size, \textbf{LM} = Facial Landmark Features, \textbf{CF} = Cross-Fusion Module, \textbf{CD} = Cycle Diffusion Strategy.
\end{tabular}

\end{table*}

\begin{figure}[!htb] \centering
    \includegraphics[width=0.48\textwidth]{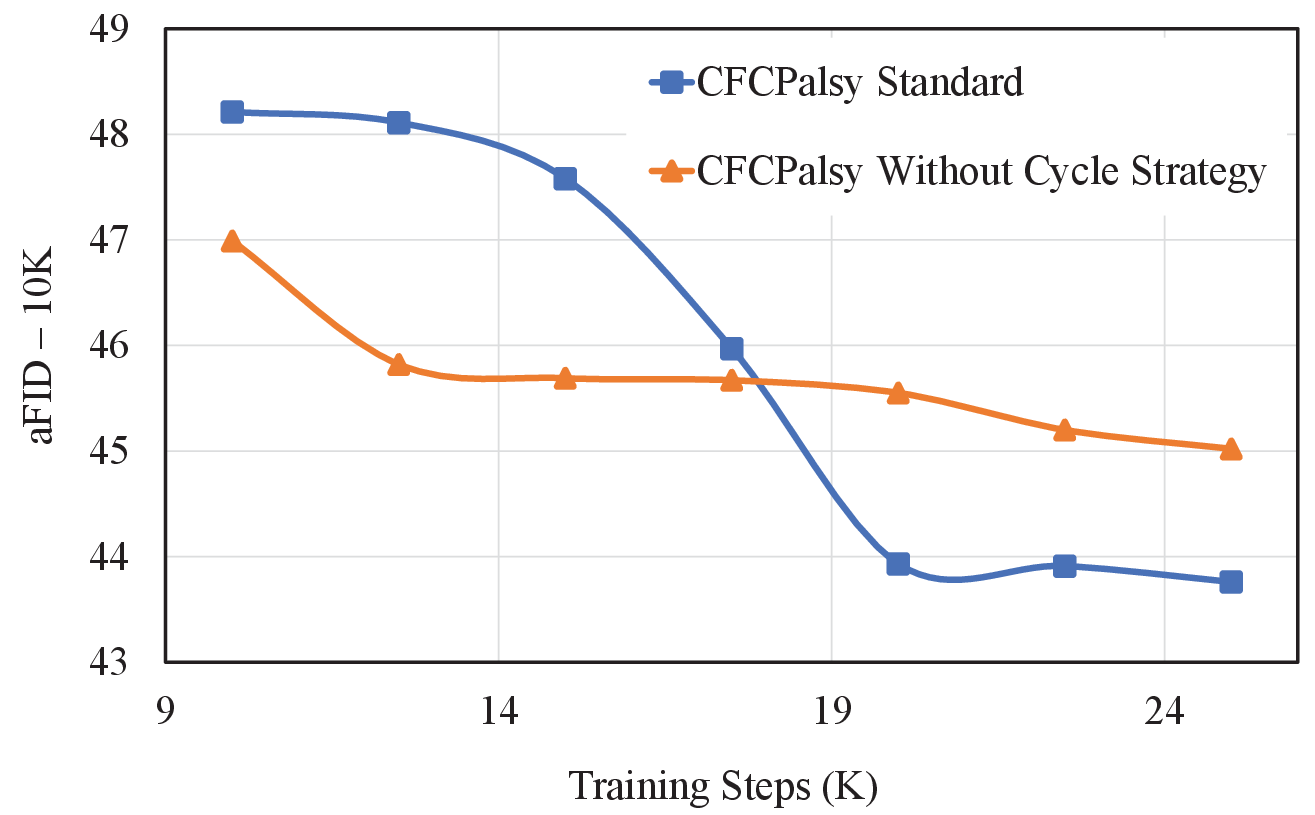}
    \caption{Ablation results for the Cycle Diffusion Training Strategy, tracking training steps to calculate aFID on the test set.} \label{fig:afid}
\end{figure}

To address this issue, we propose the aligned FID (aFID) (see Fig. \ref{fig:fid}) to evaluate the quality of the multi-condition diffusion model. Specifically, for each facial palsy patient in the test set, we use images of other individuals as identity images and the patient's condition image as the facial palsy expression image. This process synthesizes an image where the patient's facial palsy expression is transferred onto another individual. Next, we use this generated dataset as the facial palsy expression images and the original patient's images as identity images to generate a second set of synthetic images. Finally, we calculate the FID score between this generated dataset and the real dataset. Ideally, the images in these two datasets should be identical.

\subsection{Implementation details}

For the identity feature extractor $E_{ID}$, we utilized the same IR-50 \cite{deng2019arcface} as DCFace \cite{kim2023dcface}. Intermediate features from this network were injected into the cross-fusion module, and the final features were incorporated into the model prediction network with timesteps, and after initialization from DCFace, the model is trained end-to-end. For both the identity model and the face model within the expression extracting module, we also used the IR-50 architecture. After the same initialization, the ID model's weights were fixed, and the module was trained on the AffectNet \cite{mollahosseini2017affectnet} dataset, after which the module's weights were all fixed. For the landmark extractor $E_{LM}$ and detector $D_{LM}$, we used the MobileFaceNet \cite{chen2021pytorch} and fixed its weights to obtain the landmark information. For each patient's video in the MEEI dataset, we performed frame extraction. Keyframes representing eight different expressions for each facial palsy patient were manually selected. Each expression consists of 32 frames, capturing the full expression process from onset, development, peak, dissipation, to a neutral state. For the images in AFLFP and MEEI, we first performed face extraction and segmented them into 112x112 facial images. After removing low-quality images and those with poor segmentation quality, we ultimately obtained 91,168 facial images from 313 individuals. We used 88\% of the data (80,288 images from 276 individuals) for the model's training set, while 12\% of the data (10,880 images from 37 individuals) was set aside as the test set for subsequent evaluation metric calculations. CFCPalsy was trained on the training set using the AdamW Optimizer 
\cite{kingma2014adam, loshchilov2017decoupled} 
for 10 epochs with a learning rate of 0.0001, utilizing two NVIDIA RTX 4090 GPUs.

\subsection{Ablation Study}
To evaluate the contribution of each component in our model, we conducted a series of ablation studies. These experiments assess the impact of removing or altering key modules on the model's performance. The visualization results of the ablation study are shown in Fig \ref{fig:vis}. The quantitative results are shown in Tab. \ref{tab:metric}. 

\paragraph{Facial Landmarks}
To investigate the role of facial landmark features in image generation, we remove the facial landmark condition from CFCPalsy. Without learning landmark features, the model relies solely on identity and expression features in the fusion module, using query-key interactions and concatenation for information exchange. Additionally, the landmark loss is excluded from weight updates during training. The results in Tab. \ref{tab:metric} show a significant increase in aFID, while pixel-level comparison metrics for facial palsy expressions also decline notably. As illustrated in Fig. \ref{fig:vis}, the model fails to accurately learn facial palsy expressions, underscoring the critical role of facial landmark features in synthesizing detailed facial palsy expressions.

\paragraph{Cross-Fusion Module}
Tab. \ref{tab:metric} compares the performance of the model with and without the cross-fusion module. It can be observed that directly concatenating multiple features and injecting them into the model leads to a sharp increase in aFID by 10.66, making it the component with the most significant impact on overall performance. Combined with other metrics, this shows that the cross-fusion module plays a crucial role in improving both identity preservation and facial palsy feature transfer. This feature fusion module effectively enhances the model's ability to learn from multiple features comprehensively.

\paragraph{Cycle Training Strategy}
The Cycle Diffusion strategy is designed to enhance training efficiency on limited datasets. To validate its effectiveness, we conducted comparative experiments. As shown in Tab. \ref{tab:metric}, the cycle training strategy significantly improves the model's performance. Additionally, we tracked the training process and plotted the changes in aFID, as illustrated in Fig. \ref{fig:afid}. Compared to the standard diffusion model training strategy, Cycle Diffusion demonstrates superior training efficiency.

\subsection{Comparison with Previous Methods}
\label{sub:comp}
We compared CFCPalsy with DCFace \cite{kim2023dcface} on a facial synthesis task using 3x3 and 5x5 grid sizes on unseen images. The visual and numerical comparison results are presented in Fig. \ref{fig:vis} and Tab. \ref{tab:metric}, respectively. While DCFace 5x5 achieves better scores on certain metrics for generated and facial palsy expression data, these scores do not necessarily reflect superior facial palsy feature transfer. Pixel-level and perception-based metrics also account for factors like lighting and background, which are not the primary focus of facial palsy synthesis. Compared to DCFace 5x5, CFCPalsy achieves a ~22\% reduction in aFID and demonstrates a notable advantage in identity feature preservation.

\section{Conclusion}
\label{sec:Conclusion}
Our research proposes Cross-Fusion Cycle Palsy Expression Generative Model (CFCPalsy) , based on the conditional diffusion model, utilizing multiple feature extractors to accurately capture various facial features, employing a cross-fusion module to integrate these features and applying a cycle training strategy to improve training efficiency and effectiveness. As a result, CFCPalsy generates realistic facial palsy expression images, providing valuable training data for machine learning systems used in facial palsy diagnosis and treatment assistance. Results indicates that CFCPalsy surpasses existing methods in several metrics, including aFID, PSNR, and SSIM, achieving excellent performance. 

\section*{Acknowledgment}
\label{sec:Acknowledgment}
This work was supported by Shandong Provincial Natural Science Foundation under Grant ZR2024QF018.



{\small
\bibliographystyle{IEEEtran}
\bibliography{conference_101719}
}

\end{document}